%% file: 2021_xacro_misunderstandings.tex
\documentclass[hidelinks,conference,letterpaper]{IEEEtran}
\IEEEoverridecommandlockouts
\usepackage{cite}
\usepackage{amsmath,amssymb,amsfonts}
\usepackage{algorithmic}
\usepackage{graphicx}
\usepackage{textcomp}
\usepackage{xcolor}
\usepackage{listings}
\usepackage{color}
\usepackage{url}
\usepackage{comment}
\usepackage{fancyvrb} 
\usepackage{hyperref}
\usepackage{academicons}
\usepackage{xspace}
\usepackage{flushend}
\usepackage[caption=false,font=normalsize,labelfont=sf,textfont=sf]{subfig}
\usepackage{dblfloatfix}
\usepackage{titling}
\graphicspath{{figures/}}

\def\orcid#1{\kern .08em\href{https://orcid.org/#1}{\includegraphics[keepaspectratio,width=0.7em]{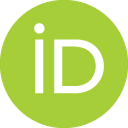}}}

\input{latex_xml_style}

\newcommand\xacro{Xacro\xspace}
\newcommand\xacros{Xacros\xspace}
\newcommand\xacroprogram{\texttt{xacro}\xspace}
\newcommand\python{\texttt{Python}\xspace}
\newcommand\urdf{URDF\xspace}

\newcommand\rviz{{\sc Rviz}\xspace}
\newcommand\gazebo{{\sc Gazebo}\xspace}
\newcommand\rosanswers{\href{https://answers.ros.org/}{answers.ros.org}\xspace}
\newcommand\categoryspace{\vspace{0.25cm}}

\newcommand\RQ[2][\textbf]{#1{RQ$_{#2}$}\xspace}

\def\BibTeX{{\rm B\kern-.05em{\sc i\kern-.025em b}\kern-.08em
    T\kern-.1667em\lower.7ex\hbox{E}\kern-.125emX}}
\begin{document}

\title{Understanding Xacro Misunderstandings}

\author{Nicholas Albergo, Vivek Rathi, and John-Paul Ore
\thanks{*This work is supported by NSF-NRI-USDA-NIFA 2021-67021-33451 and North Carolina State University Funds.}
\thanks{Nicholas Albergo and John-Paul Ore are with the Department of Computer Science, North Carolina State University, Raleigh, NC 27695, USA (email: {\tt\small \{njalberg, jwore\}@ncsu.edu})}%
\thanks{Vivek Rathi is with the Electrical and Computer Engineering Department, North Carolina State University, Raleigh, NC 27695, USA (email: {\tt\small vorathi@ncsu.edu})}}

\maketitle

\begin{abstract}
The Xacro XML macro language can be used to augment the Universal Robot Description Format (URDF) and is part of a critical toolchain from geometric representations to simulation, visualization, and system execution. However, members of the robotics community, especially newcomers, struggle to troubleshoot and understand the interplay between systems and the Xacro preprocessing pipeline.  To better understand how system developers struggle with Xacros, we manually examine 712 Xacro-related questions from the question and answer site \rosanswers and find Xacro misunderstandings fit into eight key categories using a systematic, qualitative approach called Open Coding. By examining the `tags' applied to questions, we further find that Xacro problems manifest in a befuddlingly broad set of contexts. This hinders onboarding and complicates system developers' understanding of representations and tools in the Robot Operating System. We aim to provide an empirical grounding that identifies and prioritizes impediments to users of open robotics systems, so that tool designers, teachers, and robotics practitioners can devise ways of improving robot software tooling and education. 

\end{abstract}


\vspace{0.25cm}
\section{Introduction}  

\begin{figure}[htbp]
\centerline{\includegraphics[width=7cm]{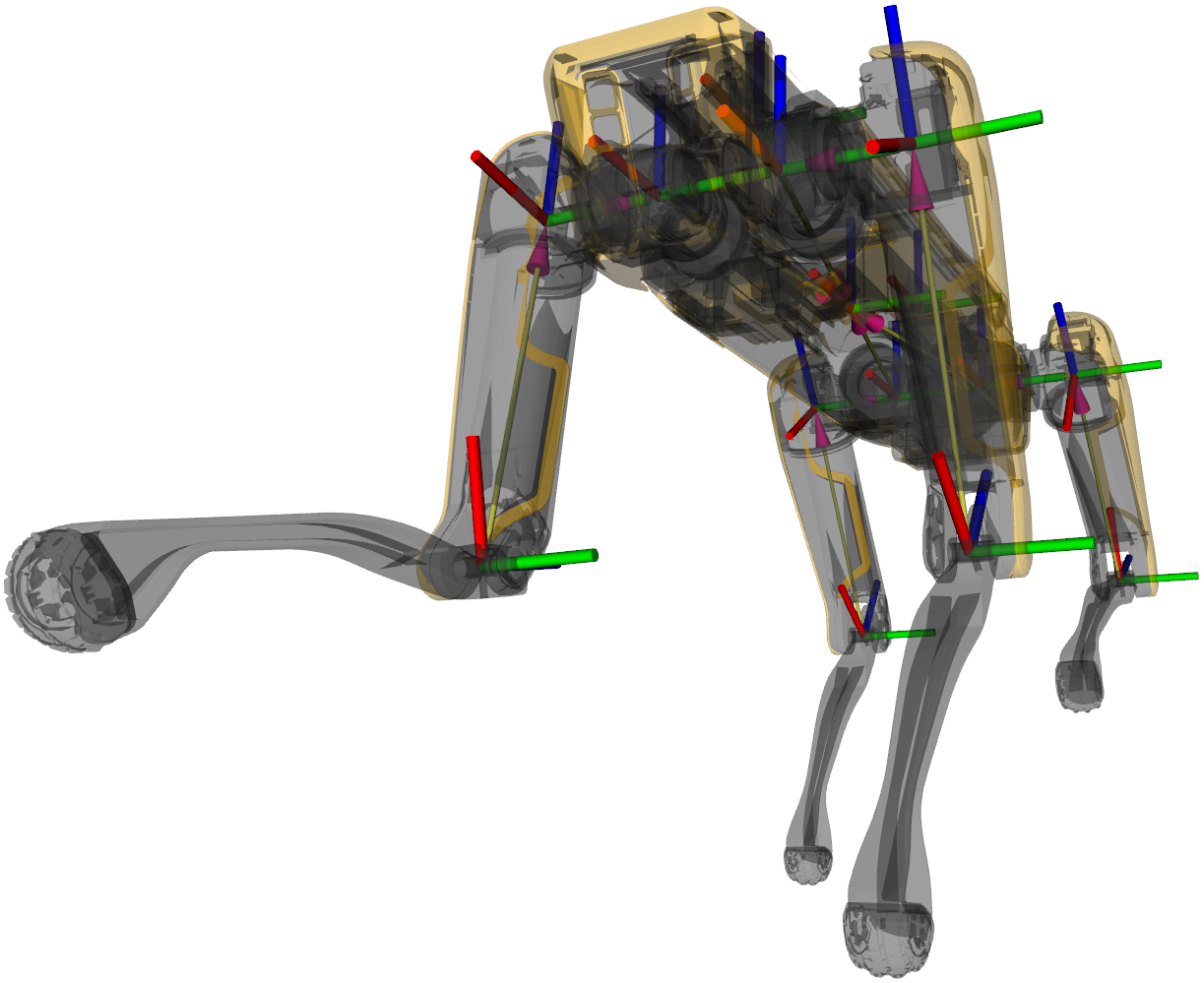}}
\caption{\emph{Geometric representation} of \emph{Spot} (Boston Dynamics) as encoded in its \xacro model (ClearPath). source: \url{https://git.io/Ju5EV}}
\label{fig:spot}
\end{figure}

A robot's \emph{geometric representation}~\cite{siciliano2008springer} is a foundational building block required by nearly every robotic system.
One format that encodes a robot's geometric representation is the Unified Robot Description Format (URDF), an XML-based encoding that encompasses reference frames, spatial kinematic chains (akin to Denavit-Hartenberg parameters~\cite{denavit1955kinematic}), linkages~\cite{hartenberg1964kinematic}, as well as the robot's visual appearance, among others. 
In the Robot Operating System (ROS)~\cite{quigley2009ros}, a URDF file is commonly augmented with commands from the Xacro\footnote{\emph{Xacro} is a portmanteau of ``XML" and ``macros.''} XML Macro Language~\cite{xacros-wiki}, and such augmented URDF files are called Xacro files.
For example, Fig.~\ref{fig:spot} shows a visual representation and reference frames of a simulated \texttt{Spot} robot from Boston Dynamics based on the \xacro model from Clearpath Robotics~\cite{clearpath-spot-github-2021}.
\xacro commands enable an expressive power in a robot model for conditional logic, variables, and leveraging physical symmetries, such as the  left-right symmetry of \texttt{Spot} shown in Fig.~\ref{fig:spot}.

The problem is that the expressive power of \xacros incurs a steeper learning curve, and the geometric representations encoded in \xacros are necessary to launch most ROS systems, visualize data (\rviz), or run simulations (\gazebo).
Newcomers to ROS might want to modify existing systems with new hardware or create entirely new systems.
When facing this learning curve, newcomers to the ROS ecosystem utilize a variety of help-seeking behaviors, including high-quality, human-driven Q\&A forums~\cite{DBLP:conf/chi/HarperRRK08} that remain an enduring and popular way for the ROS community to share knowledge and identify domain-specific authorities~\cite{jurczyk2007discovering}.

Therefore, we seek to better understand \xacro misunderstandings to help ROS designers, educators, tool builders, and system developers understand the barriers to newcomers to robotic software because barriers to onboarding impacts the long-term viability of open-source ecosystems~\cite{DBLP:conf/icsm/KolakAGHT20}.
Therefore, in this work we investigate the following research questions:

\vspace{0.1cm}

\noindent \RQ{1}. What are the main categories of misunderstandings about \xacros?

\noindent \RQ{2}. In what contexts do \xacro misunderstandings usually occur?

\vspace{0.1cm}

To address \RQ{1}, we examine a dataset of 712 questions asked on the question and answer site \rosanswers, which is used extensively by the ROS community. 
We perform \emph{Open Coding}~\cite{ezzy2013qualitative,creswell2016qualitative}, which has been used to study Ardupilot software bugs~\cite{10.1145/3468264.3468559} and extended reality bugs~\cite{li2020exploratory}. 
Here we categorize and summarize the misunderstandings robot system developers have while using \xacros.

To address \RQ{2}, we examine the question `tags' that users self-assign to \xacro related questions and find that \xacro questions co-occur with many aspects of robotic systems,  meaning that the circumstances under which \xacro related issues arise do little to isolate the issue, since other cross-cutting concerns are only evident upon an examination of the \xacro related questions.

Our contributions include:
\begin{itemize}
    \item A systematic examination of 712 \xacro related questions, resulting in eight high-level categories of \xacro misunderstandings and a hierarchy to organize them.
    \item Discovering that the primary \xacro misunderstandings are lack of awareness of the \xacro processing pipeline, adding or configuring new hardware elements, or how the \xacro language works.
    \item Identifying how \xacro problems arise in a very broad range of contexts,  indicating an opportunity for Q\&A site editors and educators.
    \item A discussion of key opportunities for automated software tool builders to help developers mitigate \xacro confusion.
    \item Sharing a digital artifact containing all questions and our categorization of each one.
\end{itemize}

\section{Background}
\label{sec:background}

Fig.~\ref{fig:spot} shows a rendering of \texttt{Spot} derived from the URDF representation augmented with the XML Macro language (hereafter, `\xacro language').
A snippet of \texttt{Spot}'s \xacro file is shown in Listing~\ref{lst:xacro}.
In lines 15, 17, and 18, there are \verb+<xacro:>+ tags that utilize a conditional to optionally include a Velodyne laser into the model.

\begin{lstlisting}[language=XML,caption={Snippet of URDF used for \emph{Spot} showing use of \xacros.},label={lst:xacro}]
<robot name="spot" xmlns:xacro="http://www.ros.org/wiki/xacro">
  <link name="body">
    <visual>
      <geometry>
        <mesh filename="package://spot_description/meshes/body.dae" />
      </geometry>
    </visual>
    <collision>
      <geometry>
        <mesh filename="package://spot_description/meshes/body_collision.stl" />
      </geometry>
    </collision>
  </link>
  ...
    <xacro:if value="$(optenv SPOT_VELODYNE 0)">
   <!-- Use the Velodyne macro for the actual sensor -->
   <xacro:include filename="$(find velodyne_description)/urdf/VLP-16.urdf.xacro" />
   <xacro:VLP-16 parent="lidar_mount">
    <origin xyz="$(optenv SPOT_VELODYNE_XYZ 0 0 0)" rpy="$(optenv SPOT_VELODYNE_RPY 0 0 0)" />
   </xacro:VLP-16>
\end{lstlisting}

Developers augment URDF models with Xacros tags with the goal of making the URDF configurable, responsive to external parameters (like the environment variable \texttt{SPOT\_VELODYNE}, as shown in Listing~\ref{lst:xacro}), and to simplify maintainability by reducing redundancy. 

\begin{figure}
\centerline{\includegraphics[width=7cm]{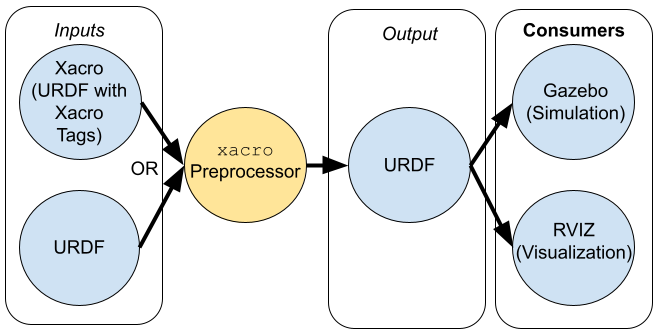}}
\caption{\xacro pipeline showing robot model inputs and simulation / visualization consumers. Inputting a URDF file without \xacro tags to the \xacro program results in the same URDF file.}
\label{fig:pipeline}
\end{figure}

Fig.~\ref{fig:pipeline} shows the \xacro processing pipeline from \xacro to \urdf.  \xacro can refer to multiple things in Fig.~\ref{fig:pipeline}, including:
\begin{itemize}
    \item \verb+<xacro:>+ tags used inside URDF files.
    \item A URDF file with the `\xacro' file suffix (i.e. `\texttt{spot.xacro}'), indicating a URDF containing \xacro tags, which is sometimes used synonymously with the robot model itself.
    \item The \xacroprogram preprocessor executable program that interprets and dynamically creates a final URDF at robot launch time.
\end{itemize}
The \xacro format consists of XML along with a subset of embedded \python paired with a preprocessor to apply transformations from \xacro-tagged URDF to URDF. 
These transformations are designed to expand, repeat, and parameterize elements to avoid redundant expressions of angular and linear relationships, which are often repeated or mirrored. 

Fig.~\ref{fig:spot} shows how \texttt{Spot} has left-right symmetry about a vertical plane bisecting the length of \texttt{Spot's} body. The link lengths, joint limits, and kinematic chain from the body to the `paws' is identical except for orientation.

Fig.~\ref{fig:pipeline} depicts the inputs to the \xacro preprocessor, showing how URDF, possibly augmented with \xacro commands, are transformed into a transient URDF consumed by \gazebo and \rviz.
\xacros also support conditional logic, variables, parameters, and inline \texttt{python} code~\cite{xacros-wiki}.

\section{Methodology}
\label{sec:methodology}

\begin{figure*}[h]
\centerline{\includegraphics[width=17.75cm]{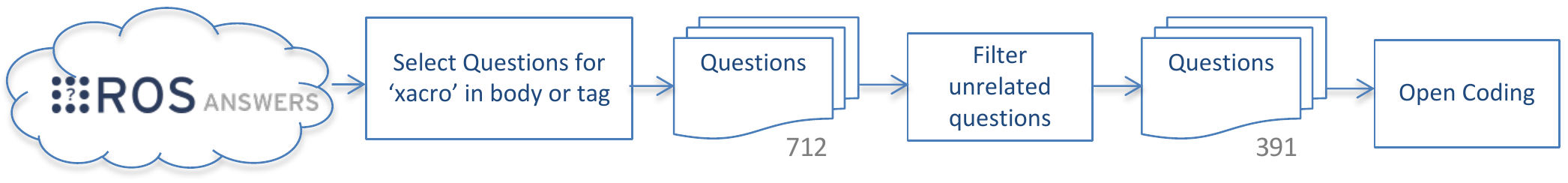}}
\caption{Process used to construct the question and answer corpus used in this work.}
\label{fig:corpus}
\end{figure*}

To collect and examine \xacro-related questions, we follow the process shown in Fig.~\ref{fig:corpus}.
In the figure, we start by using the built-in search capabilities of \rosanswers with the keyword `xacro' (case-insensitive).
Note that this yields 1)~questions with \xacro anywhere in the question or answer(s) and 2)~questions with `xacro' tags added by either the original poster or an editor.
The dates of the questions we examined ranged from February 2011 to May 2021.
This initial search finds 991 questions, of which we only consider the newest 712 because of time constraints.

Continuing with Fig.~\ref{fig:corpus}, we then examine all 712 and determine if the question pertains to \xacros, since some questions have the word `xacro' somewhere on the page but it is unrelated, i.e. the question is about material rendering, but `xacro' is in the name of the file containing the rendering. 
Filtering irrelevant questions leaves 391 questions pertaining to \xacros that we use for the Open Coding. 
The full list of questions and our categorization are available at \url{https://doi.org/10.5281/zenodo.6321341}.

Following the procedure for Open Coding~\cite{ezzy2013qualitative,creswell2016qualitative}, two authors familiar with ROS independently review all questions and assign each one to a category, \emph{without determining the categories beforehand}, so that the categories `emerge' from the data itself.
After both authors complete their first pass over all questions, they make a second pass over all questions, reevaluating how questions are assigned to categories that emerged from the first pass, and consolidating or splitting categories if necessary.
Then, all three compare categories and discuss individual questions until they converge on a set of categories and all questions are assigned to a category.
Finally, one author organizes the categories into a hierarchy.

\section{Results}

\begin{figure}
\centerline{\includegraphics[width=9cm]{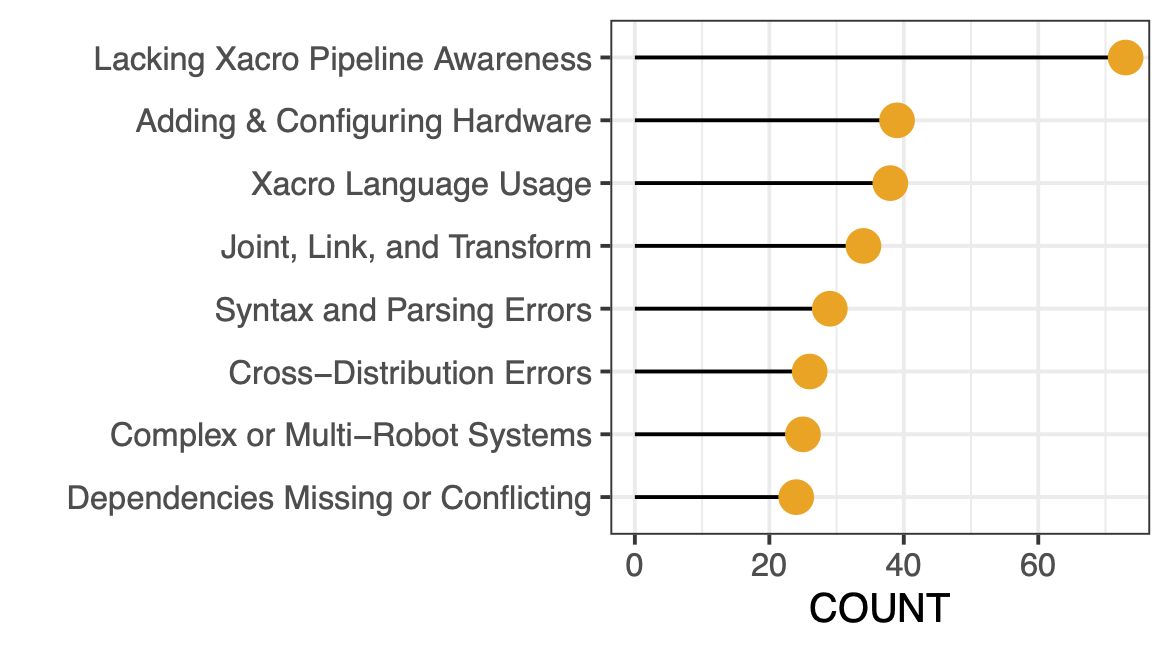}}
\caption{Distribution of \xacro questions into categories}
\label{fig:lollipop-chart}
\end{figure}

In response to \RQ{1}, our analysis yields eight distinct categories from the 391 questions considered. 
Fig.~\ref{fig:lollipop-chart} shows the counts of questions by category. As shown in the figure, lacking awareness of the \xacro pipeline is the most common misunderstanding, followed by being unsure how to add or modify new hardware.

\begin{figure}
\centerline{\includegraphics[width=9cm]{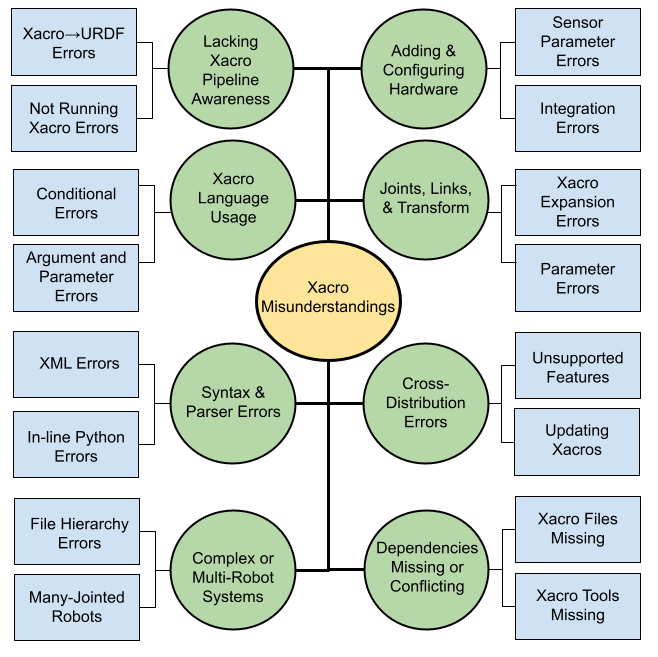}}
\caption{Hierarchical categorization of \xacro misunderstandings.}
\label{fig:hierarchy}
\end{figure}

Fig.~\ref{fig:hierarchy} shows a hierarchical view of \xacro misunderstandings, with the eight top-level categories further subdivided into the most common subcategories.  The figure shows how \xacro misunderstandings cut across other concerns like syntax, dependency bugs, incorrect paths, adding new hardware, or when the ROS distribution changes. 
The hierarchy in Fig.~\ref{fig:hierarchy} shows how the challenges of evolving systems intersect with the tools and formats used for geometric representation.

We now describe each category and include links to example questions for that category.
The full list of questions, copies of the downloaded pages (\texttt{wget}), and their assignments to categories is available in our artifact available at \url{https://doi.org/10.5281/zenodo.6321341}

\subsection{\RQ{1} Categories of \xacro Misunderstandings}

\noindent \textbf{Lacking \xacro Pipeline Awareness $19\%~(73/391)$} 
\label{sec:results:pipeline}
The largest category of \xacro misunderstanding is lacking awareness of the \xacro pipeline, as depicted in Fig.~\ref{fig:pipeline}.
Questions in this category miss one of several key facts:
\begin{enumerate}
    \item A \urdf file is created dynamically by the \xacroprogram program.
    \item Modifying a \xacro-generated \urdf file will not change the original \xacro file.
    \item The \xacroprogram program is different than a \urdf file with \xacro tags.
    \item The URDF created by the \xacroprogram program is often transient.
\end{enumerate}

Several flavors of this category exist, such as not being aware of \xacro entirely, not knowing how to use certain features that \xacro provides, or not understanding the structure of the pipeline. For instance, some users may not understand the way that \gazebo and \rviz consume the \urdf produced by the \xacroprogram program.
They may believe the \xacro file is consumed directly by \gazebo instead of being first converted by the \xacroprogram program to a URDF (see \S~\ref{sec:background}).

Example: \url{https://bit.ly/3lbC50c}~~This example shows a user editing a \xacro-generated \urdf instead of editing the original \xacro file. The changes to the \urdf do not propagate back to the \xacro file.

\categoryspace
\noindent \textbf{Adding and Configuring Hardware $10\%~(39/391)$.}
Users who seek to integrate new components into an existing model can either create a new \xacro component from scratch, copy-and-paste into their \xacro file, or import an external \xacro into their existing \xacro model. 
This was the most common system maintenance or evolution task involving \xacros.
Being able to reconfigure hardware components in a simulated robotic environment is critical to running tests.
Some configurable systems, such as the Universal Robotics UR series, can be used with a variety of additional sensors or end effectors.
ROS Industrial (\url{https://rosindustrial.org/}) provides resources in this domain, but users often face issues with obtaining the parameters for the hardware as well as a lack of knowledge about how to add the component to the robot model.
Other issues stem from integrating sensors such as cameras, Lidar, or sonar, which may have custom parameters depending on how the hardware is modeled in \gazebo.
Moreover, misconfiguring the \xacro files so that sensors publish on an incorrect topic name or namespace is also a prevalent issue faced by less experienced developers, especially with modular or deeply nested \xacro files.

Example: \url{https://bit.ly/3lhmHPY}~~The example provides a case where a user struggles to add a gripper to a robotic manipulator.

\categoryspace
\noindent \textbf{\xacro Language Usage $10\%~(38/391)$.}
This category encompasses questions that reflect confusion about arguments, parameters, and logic. 
Several questions are related to \xacro recursion (a \xacro inside of a \xacro).  Questions about arguments include how information is passed into a \xacro from a launch file or through environment variables. 
Questions about parameters reflect misunderstandings of how the \xacro language works or how and when parameters are populated during \xacro macro expansion. Conditional troubleshooting has to do with using \texttt{if} statements inside \xacro. This could be for file checking or for performing programmatic flow control inside a \xacro file.

Example: \url{https://bit.ly/3C3QmTs}~~In this example the user wants to pass arguments from a launch file to \xacro .

\categoryspace
\noindent \textbf{Joint, Link, and Transform $9\%~(34/391)$.}
Joint, link, and Transform confusions occur when a user experiences errors with how joints or links are represented by \urdf that is modified by \xacros. 
This can occur as the \xacro preprocessor assigns values to variables.
It can be hard to disentangle how joints and links are supposed to work in URDF alone from how joints and links are manipulated and modified in \xacros.  
Some users encounter these errors only when their visualization or simulation fails to render as expected.  
These questions are usually unaware of static debugging tools like \texttt{check\_urdf}.

Example: \url{https://bit.ly/3IDnRzy}~~The example shows a user struggling to understand a problem with two root links.

\categoryspace
\noindent \textbf{Syntax and Parser Errors $7\%~(29/391)$} 
\label{sec:results:syntax}
Syntax and parser errors happen for a variety of reasons, including when the \xacro file contains unknown tokens (typos) in the \xacro file, launch file, or \texttt{config.yaml} files; copy/paste errors; physics errors; or when specifying an incorrect path. 
Parsing errors usually occur when the \xacro file fails to pass the XML parser, and the \xacro tools throws an exception.
In this case, users can misunderstand what happened, since command-line error message can be challenging for end-users to understand. 
Syntax errors can involve not knowing how in-line operators such as \texttt{"\$\{...\}"} are evaluated by the \xacroprogram program.
Another trivial, but not uncommon syntax error is misspelling \xacro as ``xarco." 

Example: \url{https://bit.ly/3k4zoOE}~~This involves a missing \verb+<robot>+ tag in a \xacro file resulting in a \gazebo error.

\categoryspace
\noindent \textbf{Cross-Distribution Errors $7\%~(26/391)$.}
These include problems related to \xacro features that change over time and are usually associated with features released as part of a new ROS distribution (like ``Jade," ``Noetic," or ``Galactic").
The category also covers questions regarding converting a \xacro from one ROS distribution to another or ROS2.
We observe that the software repository \url{https://github.com/ros/xacro} has had 469 commits since 2013 and has evolved with many new features and improvements, not all of which are backward compatible across ROS distributions. 
Evolving ROS distributions coupled with new \xacro features can lead to errors.

Example: \url{https://bit.ly/3pvZNXX}~~The example shows an error that is resolved by updating the launch file to invoke \xacroprogram with \texttt{--inorder}, a convention introduced with the release of a new ROS distribution.

\begin{figure*}[ht!]
\centerline{\includegraphics[width=\textwidth]{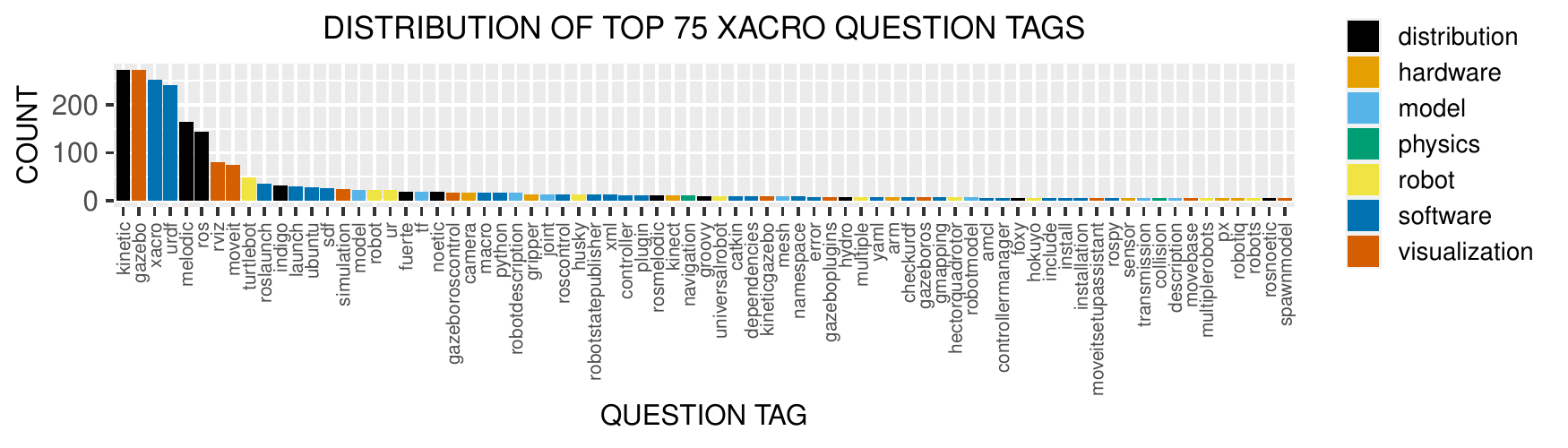}}
\caption{The first 75 of 956 different tags applied to the 391 \xacro questions examined in our study,  showing how \xacros are perceived to be intertangled with a broad set of issues.
Tags are usually applied by the question's author. }
\label{fig:answer-tags}
\end{figure*}

\categoryspace
\noindent \textbf{Complex or Multi-Robot Systems $6\%~(25/391)$.}
The \xacro \texttt{include} tag allows \xacros to be organized hierarchically.  
The allows \xacro elements to be nested in files and making it difficult for users to reason about how parameters might interact. 
We noticed this most with multi-robot systems and systems with a high degree of complexity (like the \texttt{PR2}~\cite{bohren2011towards}).
Correctly modifying and troubleshooting problems during \xacro processing of these files can be daunting. 
Complicated robots include those that use less common means of locomotion with many symmetries, such as a hexapod robot or ones with large quantities of links and joints. 
Multi-robot systems are challenging as they generally require \xacro files for each robot which can cause issues with namespaces, transforms, maps, and \texttt{robot\_description} files.
Part of this challenge is that namespaces and transforms are usually specific to individual systems whereas maps are shared.

Example: \url{https://bit.ly/35HgnNj}~~The example shows a user who wants to launch multiple robots but it not sure how to organize their \xacro file.

\categoryspace
\noindent \textbf{Dependencies Missing or Conflicting $6\%~(24/391)$.}
Dependency bugs~\cite{DBLP:conf/icse/Fischer-Nielsen20} are a frequent source of \xacro confusion.
When simulating robots, the situation is similar---there are a large variety of software artifacts that must be accessed simultaneously for a simulation to run successfully.
This can manifest when the \xacro program has not been installed, or a \xacro file with \texttt{xacro:include} tags does not have the correct files installed.

Example: \url{https://bit.ly/3sybeAf}~~The example involves missing dependencies after an \texttt{apt-get upgrade}.

\categoryspace
\noindent \textbf{Other and Minor Categories $10\%~(39/391)$.}
Categories that make up less than 3\% of the total questions reviewed are categorized as `other,' and do not clearly belong to one of the other categories.
This includes questions specific to in-line \python, specific robots, problems with \xacros in VMs (usually simulator or GPU-related), or interfacing \xacros with SDF. 
This category includes questions presented as a \xacro problem but are really about physics, pasting a complete \xacro as a way of giving other people a way to recreate their error, or questions asking for unsupported functionality. 
Example: \url{https://bit.ly/3lfFqvj}~~In this example the user asks about the \xacroprogram program being able to read parameters from the parameter server, which was not possible at the time the question was asked.

\subsection{\RQ{2} Contexts in which \xacro Problems Manifest }

In \RQ{2}, we use question tags as a proxy for assessing the contexts in which a \xacro question is asked.
We think this in an indication of the original poster's  understanding of what concepts (tags) are related to their question.
Tags on \rosanswers are self-assigned, so posters can add tags they feel are relevant. Tags enable the potential for better organization and aggregation of specific ideas. 
They can be a useful way to search the site for particular terms or topics of interest. 
However, the quality and relevance of the tags used on an arbitrary post depends on the user's level of expertise or experience. 
This leads to a large amount of tags that may not necessarily be useful or align with the question being asked.

To address this question, we collected all the tags on our corpus of 391 questions, and counted their frequency.  
Since the questions had been processed based on having `xacro' in the body or tag (See Fig.~\ref{fig:corpus}),  we expected to see clusters of tags that revealed issues that tend to co-occur with \xacro misunderstandings.
However, we found that questions involving \xacros span a large set of tags (956) with only a few obvious tags (like `ROS' or `xacro') applied frequently.

The results for \RQ{2} are summarized by Fig.~\ref{fig:answer-tags}, which shows the 75 most commonly applied tags related to \xacros. 
Although there are a wide variety of tags, the tags can be organized into similar types, such as being specific to a particular robot, as shown in the legend.  
The salient feature is the long tail showing that \xacro confusions co-occur with a wide breadth of concerns. 
Almost every aspect of using ROS, \rviz, and \gazebo are contained within these tags.

Of all the tags associated to \xacro-related questions, visualization and simulation are amoung the most common, as shown in Fig.~\ref{fig:answer-tags}.
Many errors with \xacros manifest when using visualization or simulation tools.
When simulations or visualizations fails to render or behave correctly, users might question whether their system is represented correctly in the \xacro file.
As previous empirical studies have shown, simulation software challenges are a key barrier to adoption by robotics practitioners~\cite{DBLP:journals/corr/abs-2004-07368}. This poses a challenge as simulators like \gazebo and visualizers like \rviz consume the \urdf created by the \xacroprogram (see \S~\ref{sec:background}).

Overall, the results of \RQ{2} support the notion that \xacro questions can arise in a wide variety of contexts and that \xacro misunderstandings do not appear in isolation, but almost always with an adjacent issue, like visualization, simulation, dependencies, launch files, and system evolution.

\section{Discussion}

Basic misunderstanding of how the \xacro processing pipeline works is most common causes of confusion for \xacros.
Nearly all \xacro-related errors manifest as text-based compiler messages, which have been shown to be difficult for developers to interpret~\cite{DBLP:conf/iticse/BeckerDPBB0KKMO19}.  
Therefore, we suggest that robot software tool builders focus on an Integrated Development Environment (IDE) extension or improvements to error messages generated by the \xacroprogram program. 
For the IDE extension, we suggest a targeted plug-in specifically for the \xacro format, linked to the \xacroprogram program on the backend.
Such an extension could statically check \xacro files and provided targeted, \xacro aware feedback while the \xacro file is being edited.
Another approach, \url{https://github.com/hauptmech/odio_urdf}, bypasses XML entirely and creates \urdf directly from a \python representation.

Alternatively, developers seeking to improve tools in the robot software ecosystem should focus on improving feedback provided by the command-line \xacroprogram preprocessor, since developers using text-based compiler tools spend $12\%-25\%$ of their debug time reading compiler errors~\cite{DBLP:conf/icse/BarikSLHFMP17}.  
The \xacro questions identified in this work and included in our artifact could be a useful starting point as an initial test set.

Practitioners or educators seeking to onboard new staff might start their discussions of URDFs and ROS with a presentation of the \xacro pipeline.
Educators might use the eight categories identified in this paper to develop a practical module on how robot geometric representation is encoded in ROS.
Additionally, asking students to modify or extend the \xacro of an existing system might demonstrate competency with robot software system maintenance and evolution.



\section{Conclusions}  

In this work, we examined the ways in which \xacros are misunderstood by categorizing 391 questions from \rosanswers and found eight distinct categories, with a lack of awareness of the \xacro processing pipeline as the most common misunderstanding.
We hope that our work provides empirical evidence to guide robot software tool developers so that \xacro problems are easier to recognize and understand.

\section*{Acknowledgments}
This work was partially supported by NSF-NRI \#2021-67021-33451. Any opinions, findings, and conclusions or recommendations expressed in this material are those of the authors and do not necessarily reflect the views of these agencies.
The authors would like to thank Samuel Hodges for rendering the \texttt{Spot} robot, and Aimee Allard for editing this manuscript.  

\bibliographystyle{IEEEtran}
\bibliography{IEEEabrv,mybibfile}

\end{document}

%% file: latex_xml_style.tex
\definecolor{dkgreen}{rgb}{0,0.6,0}
\definecolor{gray}{rgb}{0.5,0.5,0.5}
\definecolor{mauve}{rgb}{0.58,0,0.82}
\definecolor{gray}{rgb}{0.4,0.4,0.4}
\definecolor{darkblue}{rgb}{0.0,0.0,0.6}
\definecolor{lightblue}{rgb}{0.0,0.0,0.9}
\definecolor{cyan}{rgb}{0.0,0.6,0.6}
\definecolor{darkred}{rgb}{0.6,0.0,0.0}

\lstdefinelanguage{XML}
{
  morestring=[s][\color{mauve}]{"}{"},
  morestring=[s][\color{black}]{>}{<},
  morecomment=[s]{<?}{?>},
  morecomment=[s][\color{dkgreen}]{<!--}{-->},
  stringstyle=\color{black},
  identifierstyle=\color{lightblue},
  keywordstyle=\color{red},
  morekeywords={xmlns,xsi,noNamespaceSchemaLocation,type,id,x,y,source,target,version,tool,transRef,roleRef,objective,eventually}
}

%% file: 2021_xacro_misunderstandings.bbl
\begin{thebibliography}{10}
\providecommand{\url}[1]{#1}
\csname url@samestyle\endcsname
\providecommand{\newblock}{\relax}
\providecommand{\bibinfo}[2]{#2}
\providecommand{\BIBentrySTDinterwordspacing}{\spaceskip=0pt\relax}
\providecommand{\BIBentryALTinterwordstretchfactor}{4}
\providecommand{\BIBentryALTinterwordspacing}{\spaceskip=\fontdimen2\font plus
\BIBentryALTinterwordstretchfactor\fontdimen3\font minus
  \fontdimen4\font\relax}
\providecommand{\BIBforeignlanguage}[2]{{%
\expandafter\ifx\csname l@#1\endcsname\relax
\typeout{** WARNING: IEEEtran.bst: No hyphenation pattern has been}%
\typeout{** loaded for the language `#1'. Using the pattern for}%
\typeout{** the default language instead.}%
\else
\language=\csname l@#1\endcsname
\fi
#2}}
\providecommand{\BIBdecl}{\relax}
\BIBdecl

\bibitem{siciliano2008springer}
B.~Siciliano, O.~Khatib, and T.~Kr{\"o}ger, \emph{Springer handbook of
  robotics}.\hskip 1em plus 0.5em minus 0.4em\relax Springer, 2008, vol. 200.

\bibitem{denavit1955kinematic}
J.~Denavit and R.~S. Hartenberg, ``A kinematic notation for lower-pair
  mechanisms based on matrices,'' 1955.

\bibitem{hartenberg1964kinematic}
R.~Hartenberg and J.~Danavit, \emph{Kinematic synthesis of linkages}.\hskip 1em
  plus 0.5em minus 0.4em\relax New York: McGraw-Hill, 1964.

\bibitem{quigley2009ros}
M.~Quigley, K.~Conley, B.~Gerkey, J.~Faust, T.~Foote, J.~Leibs, R.~Wheeler,
  A.~Y. Ng \emph{et~al.}, ``Ros: an open-source robot operating system,'' in
  \emph{ICRA workshop on open source software}, vol.~3, no. 3.2.\hskip 1em plus
  0.5em minus 0.4em\relax Kobe, Japan, 2009, p.~5.

\bibitem{xacros-wiki}
S.~Glaser, W.~Woodall, and R.~Haschke, ``{XACRO} tool wiki page,''
  \url{http://wiki.ros.org/xacro}, 2021, [Online; accessed 12-September-2021].

\bibitem{clearpath-spot-github-2021}
C.~Robotics, ``{Spot Description},''
  \url{https://github.com/clearpathrobotics/spot_ros}, 2021, [Online; accessed
  11-September-2021].

\bibitem{DBLP:conf/chi/HarperRRK08}
\BIBentryALTinterwordspacing
F.~M. Harper, D.~R. Raban, S.~Rafaeli, and J.~A. Konstan, ``Predictors of
  answer quality in online q{\&}a sites,'' in \emph{Proceedings of the 2008
  Conference on Human Factors in Computing Systems, {CHI} 2008, 2008, Florence,
  Italy, April 5-10, 2008}, M.~Czerwinski, A.~M. Lund, and D.~S. Tan,
  Eds.\hskip 1em plus 0.5em minus 0.4em\relax {ACM}, 2008, pp. 865--874.
  [Online]. Available: \url{https://doi.org/10.1145/1357054.1357191}
\BIBentrySTDinterwordspacing

\bibitem{jurczyk2007discovering}
P.~Jurczyk and E.~Agichtein, ``Discovering authorities in question answer
  communities by using link analysis,'' in \emph{Proceedings of the sixteenth
  ACM conference on Conference on information and knowledge management}, 2007,
  pp. 919--922.

\bibitem{DBLP:conf/icsm/KolakAGHT20}
\BIBentryALTinterwordspacing
S.~Kolak, A.~Afzal, C.~L. Goues, M.~Hilton, and C.~S. Timperley, ``It takes a
  village to build a robot: An empirical study of the {ROS} ecosystem,'' in
  \emph{{IEEE} International Conference on Software Maintenance and Evolution,
  {ICSME} 2020, Adelaide, Australia, September 28 - October 2, 2020}.\hskip 1em
  plus 0.5em minus 0.4em\relax {IEEE}, 2020, pp. 430--440. [Online]. Available:
  \url{https://doi.org/10.1109/ICSME46990.2020.00048}
\BIBentrySTDinterwordspacing

\bibitem{ezzy2013qualitative}
D.~Ezzy, \emph{Qualitative analysis}.\hskip 1em plus 0.5em minus 0.4em\relax
  Routledge, 2013.

\bibitem{creswell2016qualitative}
J.~W. Creswell and C.~N. Poth, \emph{Qualitative inquiry and research design:
  Choosing among five approaches}.\hskip 1em plus 0.5em minus 0.4em\relax Sage
  publications, 2016.

\bibitem{10.1145/3468264.3468559}
\BIBentryALTinterwordspacing
D.~Wang, S.~Li, G.~Xiao, Y.~Liu, and Y.~Sui, ``An exploratory study of
  autopilot software bugs in unmanned aerial vehicles,'' in \emph{Proceedings
  of the 29th ACM Joint Meeting on European Software Engineering Conference and
  Symposium on the Foundations of Software Engineering}, ser. ESEC/FSE
  2021.\hskip 1em plus 0.5em minus 0.4em\relax New York, NY, USA: Association
  for Computing Machinery, 2021, p. 20–31. [Online]. Available:
  \url{https://doi-org.prox.lib.ncsu.edu/10.1145/3468264.3468559}
\BIBentrySTDinterwordspacing

\bibitem{li2020exploratory}
S.~Li, Y.~Wu, Y.~Liu, D.~Wang, M.~Wen, Y.~Tao, Y.~Sui, and Y.~Liu, ``An
  exploratory study of bugs in extended reality applications on the web,'' in
  \emph{2020 IEEE 31st International Symposium on Software Reliability
  Engineering (ISSRE)}.\hskip 1em plus 0.5em minus 0.4em\relax IEEE, 2020, pp.
  172--183.

\bibitem{bohren2011towards}
J.~Bohren, R.~B. Rusu, E.~G. Jones, E.~Marder-Eppstein, C.~Pantofaru, M.~Wise,
  L.~M{\"o}senlechner, W.~Meeussen, and S.~Holzer, ``Towards autonomous robotic
  butlers: Lessons learned with the pr2,'' in \emph{2011 IEEE International
  Conference on Robotics and Automation}.\hskip 1em plus 0.5em minus
  0.4em\relax IEEE, 2011, pp. 5568--5575.

\bibitem{DBLP:conf/icse/Fischer-Nielsen20}
\BIBentryALTinterwordspacing
A.~Fischer{-}Nielsen, Z.~Fu, T.~Su, and A.~Wasowski, ``The forgotten case of
  the dependency bugs: on the example of the robot operating system,'' in
  \emph{{ICSE-SEIP} 2020: 42nd International Conference on Software
  Engineering, Software Engineering in Practice, Seoul, South Korea, 27 June -
  19 July, 2020}, G.~Rothermel and D.~Bae, Eds.\hskip 1em plus 0.5em minus
  0.4em\relax {ACM}, 2020, pp. 21--30. [Online]. Available:
  \url{https://doi.org/10.1145/3377813.3381364}
\BIBentrySTDinterwordspacing

\bibitem{DBLP:journals/corr/abs-2004-07368}
\BIBentryALTinterwordspacing
A.~Afzal, D.~S. Katz, C.~L. Goues, and C.~S. Timperley, ``A study on the
  challenges of using robotics simulators for testing,'' \emph{CoRR}, vol.
  abs/2004.07368, 2020. [Online]. Available:
  \url{https://arxiv.org/abs/2004.07368}
\BIBentrySTDinterwordspacing

\bibitem{DBLP:conf/iticse/BeckerDPBB0KKMO19}
\BIBentryALTinterwordspacing
B.~A. Becker, P.~Denny, R.~Pettit, D.~Bouchard, D.~J. Bouvier, B.~Harrington,
  A.~Kamil, A.~Karkare, C.~McDonald, P.~Osera, J.~L. Pearce, and J.~Prather,
  ``Compiler error messages considered unhelpful: The landscape of text-based
  programming error message research,'' in \emph{Proceedings of the Working
  Group Reports on Innovation and Technology in Computer Science Education,
  ITiCSE 2019, Aberdeen, Scotland Uk, July 15-17, 2019}, B.~Scharlau,
  R.~McDermott, A.~Pears, and M.~Sabin, Eds.\hskip 1em plus 0.5em minus
  0.4em\relax {ACM}, 2019, pp. 177--210. [Online]. Available:
  \url{https://doi.org/10.1145/3344429.3372508}
\BIBentrySTDinterwordspacing

\bibitem{DBLP:conf/icse/BarikSLHFMP17}
\BIBentryALTinterwordspacing
T.~Barik, J.~Smith, K.~Lubick, E.~Holmes, J.~Feng, E.~R. Murphy{-}Hill, and
  C.~Parnin, ``Do developers read compiler error messages?'' in
  \emph{Proceedings of the 39th International Conference on Software
  Engineering, {ICSE} 2017, Buenos Aires, Argentina, May 20-28, 2017},
  S.~Uchitel, A.~Orso, and M.~P. Robillard, Eds.\hskip 1em plus 0.5em minus
  0.4em\relax {IEEE} / {ACM}, 2017, pp. 575--585. [Online]. Available:
  \url{https://doi.org/10.1109/ICSE.2017.59}
\BIBentrySTDinterwordspacing

\end{thebibliography}
